\documentclass[runningheads]{llncs}
\usepackage{graphicx}
\usepackage{amsmath,amssymb} 
\usepackage{color}
\usepackage[width=122mm,left=12mm,paperwidth=146mm,height=193mm,top=12mm,paperheight=217mm]{geometry}
\begin{document}
\pagestyle{headings}
\mainmatter

\newcommand{\new}[1]{{\textcolor{red}{#1}}}
\newcommand\blfootnote[1]{%
  \begingroup
  \renewcommand\thefootnote{}\footnote{#1}%
  \addtocounter{footnote}{-1}%
  \endgroup
}

\title{Keep it SMPL: Automatic Estimation of 3D\\ Human Pose and Shape
  from a Single Image} 

\titlerunning{Keep it SMPL}

\authorrunning{Bogo, Kanazawa, Lassner, Gehler, Romero, Black}

\author{Federica Bogo$^{2,*}$, Angjoo Kanazawa$^{3,*}$, Christoph Lassner$^1$, 
Peter Gehler$^{1,4}$, Javier  Romero$^1$,  Michael J. Black$^1$}


\institute{
	$^1$Max Planck Institute for Intelligent Systems, T\"ubingen,
  Germany\\
  $^2$Microsoft Research, \hspace{0.01in}
  $^3$University of Maryland, \hspace{0.01in}
  $^4$University of T\"{u}bingen\\
	\email{ febogo@microsoft.com}, \email{ kanazawa@umiacs.umd.edu}\\
  \email{ \{christoph.lassner, pgehler, jromero, black\}@tue.mpg.de}
}

\maketitle

\begin{abstract}
We describe the first method to automatically estimate the 3D pose of the human body as well as its 3D shape from a
single unconstrained image.
We estimate a full 3D mesh and show that
2D joints alone carry a surprising amount of information about body shape.
The problem is challenging because of the complexity of the human
body, articulation, occlusion, clothing, lighting, and the inherent ambiguity in inferring
3D from 2D.
To solve this, we first use a recently published CNN-based method, DeepCut, to
predict (bottom-up) the 2D body joint locations.
We then fit (top-down) a recently published statistical body shape model, called
SMPL, to the 2D joints.
We do so  by minimizing an objective function that penalizes
the error between the projected 3D model joints and detected 2D joints.
Because SMPL captures correlations in human shape across
the population, we are able to robustly fit it to very little data.
We further leverage the 3D model to prevent solutions that cause interpenetration.
We evaluate our method, SMPLify, on the Leeds Sports, HumanEva, and Human3.6M datasets,
showing superior pose accuracy with respect to the state of the art.

\blfootnote{$^*$ The first two authors contributed equally to this
  work. The work was performed at the MPI for Intelligent Systems.}

\end{abstract}

\section{Introduction}
\label{sec:intro}
\begin{figure}[t]
\centerline{
\includegraphics[width=\linewidth]{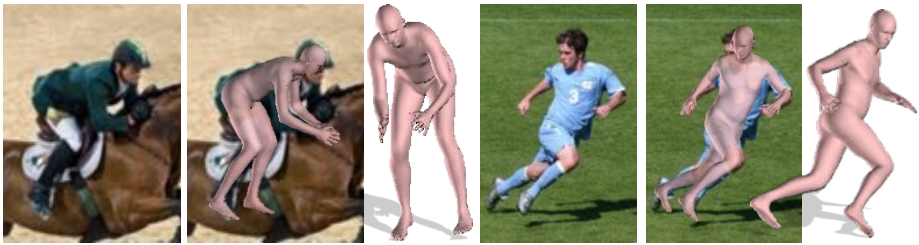}
}
\caption{{\bf Example results.} 3D pose and shape estimated by our method for two images from the Leeds Sports Pose Dataset~\cite{LSP:2010}.
We show the original image (left), our fitted model (middle), and the 3D model rendered from a different viewpoint (right).}
\label{fig:teaser}
\end{figure}

The estimation of 3D human pose from a single image
is a longstanding problem with many applications.
Most previous approaches focus only on pose and ignore 3D human shape.
Here we provide a solution that is {\em fully automatic} and estimates
a 3D mesh capturing both pose and shape from a 2D image.
We solve the problem in two steps.
First we estimate 2D joints using a recently proposed convolutional
neural network (CNN) called DeepCut \cite{Leonid2016DeepCut}.
So far CNNs have been successful at estimating 2D human pose
\cite{modeep,Pfister15a,Pfister14a,Leonid2016DeepCut,Toshev:2014} but not
3D pose and shape from one image.
Consequently we add a second step, which estimates 3D pose and shape
from the 2D joints using a 3D generative model called SMPL \cite{SMPL:2015}.
The overall framework, which we call ``SMPLify'', fits within a classical paradigm of bottom up
estimation (CNN) followed by top down verification (generative model).
A few examples are shown in  Fig.~\ref{fig:teaser}.

There is a long literature on estimating 3D pose from 2D joints.
Unlike previous methods, our approach exploits a high-quality 3D human body model that is
trained from thousands of 3D scans and hence captures the statistics of
shape variation in the population as well as how people deform with pose.
Here we use the SMPL body model \cite{SMPL:2015}.
The key insight is that such a model can be fit to very little data
because it captures so much information of human body shape.

We define an objective function and optimize pose and shape
directly, so that the projected joints of the 3D model are
close to the 2D joints estimated by the CNN.
Remarkably, fitting only 2D joints produces plausible estimates of 3D
body {\em shape}.
We perform a quantitative evaluation using synthetic data and find that 2D joint locations contain a surprising amount of 3D shape information.

In addition to capturing shape statistics, there is a second advantage
to using a generative 3D model: it enables us to reason about
interpenetration.
Most previous work in the area has estimated 3D stick figures from 2D joints.
With such models, it is easy to find poses that are impossible because
the body parts would intersect in 3D.
Such solutions are very common when inferring 3D from 2D because the
loss of depth information makes the solution ambiguous.

Computing interpenetration of a complex, non-convex, articulated object
like the body, however, is expensive.
Unlike previous work~\cite{Guan:2009,GuanThesis}, we provide an interpenetration term that is differentiable with respect to body shape and pose.
Given a 3D body shape we define a set of ``capsules'' that approximates the body shape.
Crucially, capsule dimensions are linearly regressed from model shape parameters.
This representation lets us compute interpenetration efficiently.
We show that this term helps to prevent incorrect poses.

SMPL is gender-specific; i.e. it distinguishes the shape space of females and males.
To make our method fully automatic, we introduce a gender-neutral model.
If we do not know the gender, we fit this model to images. If we know the gender, then we
use a gender-specific model for better results.

To deal with pose ambiguity, it is important to have a good pose
prior. Many recent methods learn sparse, over-complete dictionaries
from the CMU dataset~\cite{cmu} or learn dataset-specific priors.
We train a prior over pose from SMPL models that have been fit to the CMU
mocap {\em marker} data \cite{cmu} using MoSh \cite{Loper:SIGASIA:2014}.
This factors shape from pose with pose represented as relative
rotations of the body parts.
We then learn a generic multi-modal pose prior from this.

We compare the method to recently published methods~\cite{Akhter:2015,Ramakrishna:2012,Zhou:2015b} using the exact
same 2D joints as input. We show the robustness of the approach
qualitatively on images from the challenging Leeds Sports Pose Dataset (LSP)~\cite{LSP:2010} (Fig.~\ref{fig:teaser}).
We quantitatively compare the method on HumanEva-I~\cite{HumanEva:2010} and Human3.6M~\cite{Human36m:2014}, finding that our method is more accurate than previous methods.

In summary our contributions are:
1) the first fully automatic method of estimating 3D body shape and pose from 2D joints;
2) an interpenetration term that is differentiable with respect to shape and pose;
3) a novel objective function that matches a 3D body model to 2D
joints;
4) for research purposes, we provide the code, 2D joints, and 3D models for all examples in
the paper~\cite{simplifyweb}.

\section{Related Work}
\label{sec:related}
The recovery of 3D human pose from 2D is fundamentally ambiguous and all methods deal with this ambiguity in different ways.
These include user intervention, using rich image features, improving the optimization methods, and, most commonly, introducing prior knowledge.
This prior knowledge typically includes both a ``shape'' prior that
enforces anthropometric constraints on bone lengths and a ``pose''
prior that favors plausible poses and rules out impossible ones. 
While there is a large literature on estimating body pose and shape
from multi-camera images or
video sequences \cite{Balan:2007,Grest:2005,Jain:2010,Sminchisescu:2002}, here we focus on static image methods.
We also focus on methods that do not require a background image for
background subtraction, but rather infer 3D pose from 2D joints.

Most methods formulate the problem as finding a 3D {\em skeleton} such that its 3D joints project to known or estimated 2D joints.
Note that the previous work often refers to this skeleton in a
particular posture as a ``shape''.
In this work we take shape to mean the pose-invariant surface of the
human body in 3D and distinguish this from pose, which is the articulated posture of the limbs.

{\bf 3D pose from 2D joints.} 
These methods all assume known correspondence between 2D joints and a 3D skeleton. 
Methods make different assumptions about the statistics of limb-length variation. Lee and Chen \cite{Lee1985} assume known limb lengths of a stick figure  while
Taylor~\cite{Taylor:2000} assumes the ratios of limb lengths are known.
Parameswaran and Chellappa  \cite{Parameswaran:CVPR:2004} assume that
limb lengths are isometric across people, varying only in global scaling.
Barron and Kakadiaris \cite{BarronKakadiaris2001} build a statistical
model of limb-length variation from extremes taken from anthropometric tables.
Jiang \cite{Jiang} takes a non-parametric approach, treating poses in the CMU dataset~\cite{cmu} as exemplars.

Recent methods typically use the CMU dataset and learn a statistical model of limb lengths and poses from it.
For example,  both \cite{Fan:2014,Ramakrishna:2012} learn a dictionary of poses but use a fairly weak anthropometric model on limb lengths.
Akhter and Black \cite{Akhter:2015} take a similar approach but add a novel pose prior that captures pose-dependent joint angle limits.
Zhou et al. \cite{Zhou:2015b} also learn a shape dictionary but
they create a sparse basis that also captures how these poses appear from different camera views.  They show that the resulting
optimization problem is easier to solve. 
Pons-Moll et al.~\cite{Pons:2014} take a different approach:
they estimate qualitative ``posebits'' from mocap and relate these to 3D pose.

The above approaches have weak, or non-existent, models of human shape.
In contrast, we argue that a stronger model of body shape, learned from thousands of people, captures the anthropometric constraints of the population.
Such a model helps reduce ambiguity, making the problem easier. 
Also, because we have 3D shape, we can model interpenetration, avoiding impossible poses.

{\bf 3D pose and shape.} 
There is also work on estimating 3D body shape from single images.
This work often assumes good silhouettes are available.
Sigal et al.~\cite{Sigal:2008} assume that silhouettes are given, compute shape features from them, and then use a mixture of experts to predict 3D body pose and shape from the features.
Like us they view the problem as a combination of a bottom-up discriminative method and a top-down generative method.
In their case the generative model (SCAPE \cite{Anguelov:2005}) is fit to the image silhouettes.
Their claim that the method is fully automatic is only true if silhouettes are available, which is often not the case.
They show a limited set of results using perfect silhouettes and do not evaluate pose accuracy.

Guan et al.~\cite{Guan:2009,GuanThesis} take manually marked 2D joints
and first estimate the 3D pose of a stick figure using classical
methods \cite{Lee1985,Taylor:2000}.
They use the pose of this stick figure to pose a SCAPE model, project the model into the image and use this to segment the image with GrabCut~\cite{grabcut}.
They then fit the SCAPE shape and pose to a variety of features including the silhouette, image edges, and shading cues.  
They assume the camera focal length is known or approximated, the lighting is roughly initialized,  and that the height of the person is known.
They use an interpenetration term that models each body part by its convex hull. 
They then check each of the extremities to see how many other body
points fall inside it and define a penalty function that penalizes interpenetration. 
This does not admit easy optimization.  

In similar work, Hasler et al.~\cite{Hasler:2010} fit a parametric body model to silhouettes.  
Typically, they require a known segmentation and a few manually provided correspondences.
In cases with simple backgrounds, they use four clicked points on the hands and feet to establish a rough fit and then use GrabCut to segment the person.
They demonstrate this on one image.
Zhou et al. \cite{Zhou:2010} also fit a parametric model of body shape
and pose to a cleanly segmented silhouette using significant manual intervention.
Chen et al.~\cite{Chen:ECCV:2010} fit a parametric model of body shape and
pose to manually extracted silhouettes; they do not evaluate
quantitative accuracy.

To our knowledge, no previous method estimates {\em 3D body shape} and pose directly from only {\em 2D joints}.
A priori, it may seem impossible, but given a good statistical model, our approach works surprisingly well.
This is enabled by our use of SMPL \cite{SMPL:2015}, which unlike
SCAPE, has explicit 3D joints; we fit their projection directly to 2D joints.
SMPL defines how joint locations are related to the 3D surface of the
body, enabling inference of shape from joints.
Of course this will not be perfect as a person can have the exact same limb lengths with varying weight.
SMPL, however, does not represent anatomical joints, rather it represents them as a function of the surface vertices.
This couples joints and shape during model training and means that
solving for them together is important.

{\bf Making it automatic.}
None of the methods above are automatic, most assume known correspondences, and some involve significant manual intervention.
There are, however, a few methods that try to solve the entire problem of
inferring 3D pose from a single image.

Simo-Serra et al. \cite{Simoserra:2013,Simoserra:2012} take into account that 2D part detections are unreliable and formulate a probabilistic model that estimates the 3D pose and the matches to the 2D image features together.
Wang et al. \cite{Wang:2014} use a weak model of limb lengths
\cite{Lee1985} but exploit automatically detected joints in the image
and match to them robustly using an L1 distance.
They use a sparse basis to represent poses as in other methods.

Zhou et al.~\cite{Zhou:2014} run a 2D pose detector
\cite{YangCVPR11} and then optimize 3D pose, automatically rejecting outliers.
Akhter and Black \cite{Akhter:2015} run a different 2D detector
\cite{kiefel2014} and show results for their method on a few images.
Both methods are only evaluated qualitatively. 
Yasin et al.~\cite{Yasin:CVPR:2016} take a non-parametric approach in which the detected
  2D joints are used to look up the nearest 3D poses in a mocap
  dataset.
Kostrikov and Gall~\cite{Kostrikov:2014} combine regression forests and a 3D
  pictorial model to regress 3D joints.
Ionescu et al.~\cite{Ionescu:CVPR:2014} train a method to predict 3D
pose from images by first predicting body part labels; their results on
Human3.6M are good but they do not test on complex images where
background segmentation is not available.
Kulkarni et al.~\cite{Kulkarni_2015_CVPR} use a generative model of body shape
and pose, together with a probabilistic programming framework to
estimate body pose from single images.  They deal with visually simple
images, where the person is well centered and cropped, and do not evaluate 3D pose accuracy.

Recent advances in deep learning are producing methods for estimating
2D joint positions accurately \cite{Leonid2016DeepCut,Wei:CVPR:2016}.
We use the recent DeepCut method \cite{Leonid2016DeepCut}, which gives remarkably good 2D detections.
Recent work \cite{Zhou:CVPR:2016} uses a CNN to
estimate 2D joint locations and then fit 3D pose
to these using a monocular video sequence. They do not show results for
single images.

None of these automated methods estimate 3D body shape.
Here we demonstrate a complete system that uses 2D joint detections and
fits pose and shape to them from a single image.

\section{Method}
\label{sec:method}

\begin{figure}[t]
\centerline{
\includegraphics[width=\linewidth]{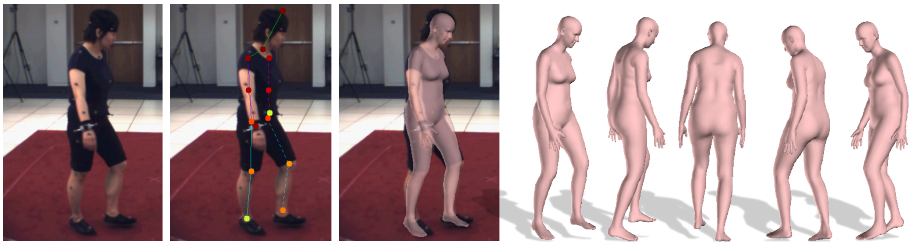}
}
\caption{{\bf System overview.} Left to right: Given a single image, we use a
  CNN-based method to predict 2D joint locations (hot colors
    denote high confidence). We then fit a 3D body model to this, to estimate 3D body shape and pose.
Here we show a fit on HumanEva~\cite{HumanEva:2010}, projected into
the image and shown from different viewpoints.}
\label{fig:overview}
\end{figure}

Figure~\ref{fig:overview} shows an overview of our system.
We take a single input image, and use the DeepCut CNN
\cite{Leonid2016DeepCut} to predict 2D body joints, $J_{\mathrm{est}}$.
For each 2D joint $i$ the CNN provides a confidence value, $w_i$.
We then fit a 3D body model such that the projected joints of the
model minimize a robust weighted error term.
In this work we use a skinned vertex-based model,
SMPL~\cite{SMPL:2015}, and call the system that takes a 2D image and produces
a posed 3D mesh, {\em SMPLify}.

The body model is defined as a function
$M(\vec{\beta},\vec{\theta},\vec{\gamma})$, parameterized by shape
${\vec{\beta}}$, pose $\vec{\theta}$, and translation $\vec{\gamma}$. The output of the function is a triangulated surface, $\mathcal{M}$, with $6890$ vertices.
Shape parameters $\vec{\beta}$ are coefficients of a low-dimensional
shape space, learned from a training set of thousands of registered
scans.
Here we use one of three shape models: male, female, and
gender-neutral. SMPL defines only male and female models.
For a fully automatic method, we trained a new gender-neutral model
using the approximately 2000 male and 2000 female body shapes used to
train the gendered SMPL models.
If the gender is known, we use the appropriate model.
The model used is indicated by its color: pink for gender-specific
and light blue for gender-neutral.

The pose of the body is defined by a skeleton rig with $23$ joints;
pose parameters $\vec{\theta}$ represent the axis-angle representation
of the relative rotation between parts.
Let $J(\vec{\beta})$ be the function that predicts 3D skeleton joint
locations from body shape.
In SMPL, joints are a sparse linear combination of surface vertices
or, equivalently, a function of the shape coefficients.
Joints can be put in arbitrary poses by applying a global rigid transformation.
In the following, we denote posed 3D joints as
$R_\theta(J(\vec{\beta})_i)$, for joint $i$, where $R_\theta$ is the global rigid
transformation induced by pose $\vec{\theta}$. SMPL defines pose-dependent deformations; for the gender-neutral shape model, we use the female deformations, which are general enough in practice.
Note that the SMPL model and DeepCut skeleton have slightly different joints.
We associate DeepCut joints with the most similar SMPL joints.
To project SMPL joints into the image we use a perspective camera model, defined by parameters $K$.

\subsection{Approximating Bodies with Capsules}
\label{subsec:capsules}
\begin{figure}[t]
\centerline{
\includegraphics[width=\linewidth]{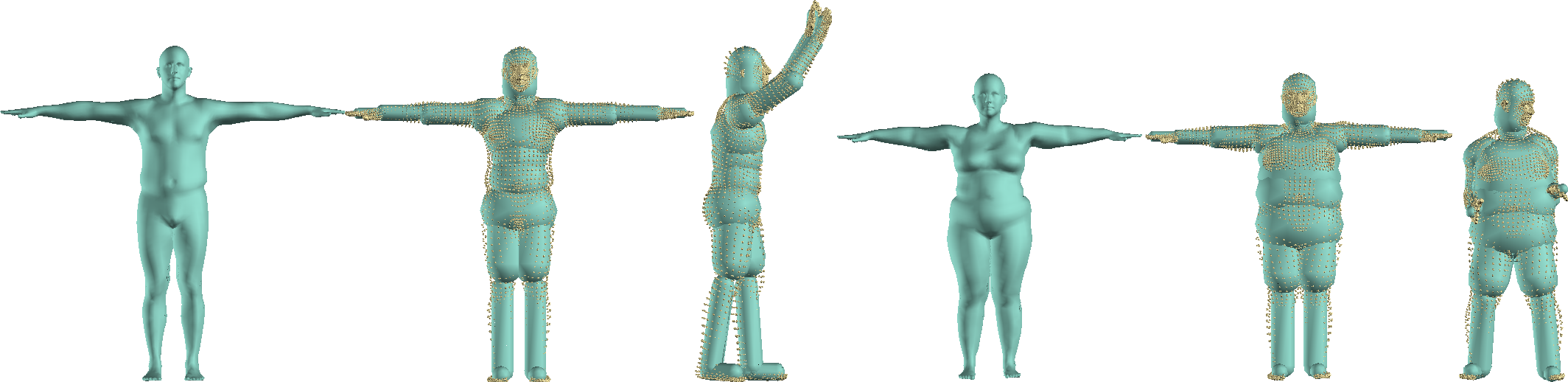}
}
\caption{{\bf Body shape approximation with capsules.}
Shown for two subjects.  Left to right: original shape, shape approximated with capsules, capsules reposed.
Yellow point clouds represent actual vertices of the model that is approximated.
}
\label{fig:capsules}
\end{figure}

We find that previous methods produce 3D poses that are impossible due
to interpenetration between body parts.
An advantage of our 3D shape model is that  it allows us to detect and prevent this.
Computing interpenetration however is expensive for complex, non-convex, surfaces like the body.
In graphics it is common to use proxy geometries to compute collisions
efficiently~\cite{collision,Thiery:SIGASIA:2013}. We follow this
approach and approximate the body surface as a set of ``capsules'' (Fig. ~\ref{fig:capsules}).
Each capsule has a radius and an axis length.

We train a regressor from model shape parameters to capsule parameters (axis length and radius), and pose the capsules
according to $R_\theta$, the rotation induced by the kinematic chain.
Specifically, we first fit $20$ capsules, one per body part, excluding fingers and toes, to
the body surface of the unposed training body shapes used to learn SMPL \cite{SMPL:2015}.
Starting from capsules manually attached to body joints in the
template, we perform gradient-based optimization of their radii and
axis lengths to minimize the bidirectional distance between capsules and
body surface.
We then learn a linear regressor
from body shape coefficients, $\vec{\beta}$, to the capsules' radii
and axis lengths using cross-validated ridge regression.
Once the regressor is trained, the procedure is iterated once more, initializing
the capsules with the regressor output.
While previous work uses approximations to detect interpenetrations~\cite{pons2015,sminchisescu2001}, we believe this regression from shape parameters is novel.

\subsection{Objective Function}
\label{subsec:objectives}

To fit the 3D pose and shape to the CNN-detected 2D joints, we minimize an objective function that is the sum of five error terms: a
joint-based data term, three pose priors, and a shape prior; that is $E(\vec{\beta}, \vec{\theta}) = $
\begin{equation}
  \label{eq:objective}
E_J(\vec{\beta}, \vec{\theta}; K, J_{\mathrm{est}}) + \lambda_{\theta} E_{\theta}(\vec{\theta}) + \lambda_a E_a (\vec{\theta}) + {\lambda_{sp}}
E_{sp}(\vec{\theta}; \vec{\beta}) + \lambda_\beta E_\beta(\vec{\beta})
\end{equation}
where $K$ are camera parameters
and $\lambda_{\theta}$, $\lambda_a$, $\lambda_{sp}$
$\lambda_\beta$ are scalar weights.

Our joint-based data term penalizes the weighted 2D distance between
estimated joints, $J_{\mathrm{est}}$, and corresponding projected SMPL joints:
\begin{align}
\label{eq:joints}
E_J(\vec{\beta}, \vec{\theta}; K, J_{\mathrm{\mathrm{est}}}) = \sum_{\mathrm{joint}\,i}  w_i \rho(\Pi_{K}(R_\theta(J(\vec{\beta})_i)) - J_{\mathrm{est},i})
\end{align}
where $\Pi_K$ is the projection from 3D to 2D induced by a camera with parameters $K$.
We weight the contribution of each joint by the confidence of its
estimate, $w_i$, provided by the CNN. For occluded joints, this value is usually low; pose in this case is driven by our pose priors.
To deal with noisy estimates, we use a robust differentiable
Geman-McClure penalty function, $\rho$, \cite{geman}.

We introduce a pose prior penalizing elbows and knees that bend unnaturally:
\begin{align}
\label{eq:angles}
E_a (\vec{\theta}) = \sum_{i}  \mathrm{exp}{(\vec{\theta}_i)},
\end{align}
where $i$ sums over pose parameters (rotations) corresponding to the
bending of knees and elbows. The exponential strongly penalizes rotations violating natural
constraints (e.g. elbow and knee hyperextending).
Note that when the joint is not bent, $\theta_i$ is zero.
Negative bending is natural and is not penalized heavily while
positive bending is unnatural and is penalized more.

Most methods for 3D pose estimation use some sort of pose prior to
favor probable poses over improbable ones.
Like many previous methods we train our pose prior using the CMU dataset~\cite{cmu}.
Given that poses vary significantly, it is important to represent the
multi-modal nature of the data, yet also keep the prior
computationally tractable.
To build a prior, we use poses obtained by fitting SMPL to the CMU marker data using MoSh~\cite{Loper:SIGASIA:2014}.
We then fit  a mixture of Gaussians to approximately
1 million poses, spanning 100 subjects. Using the mixture model directly in our optimization framework is problematic computationally because we need to optimize the negative
logarithm of a sum.
As described in~\cite{Olson:IJRR:2013}, we approximate the sum in the mixture of
Gaussians by a max operator:
\begin{align}
  E_\theta(\vec{\theta}) \equiv -\log\sum_j(g_j \mathcal{N}(\vec{\theta};\vec{\mu}_{\theta,j}, \Sigma_{\theta,j}))
             &\approx -\log (\max_j(c g_j \mathcal{N}(\vec{\theta};\vec{\mu}_{\theta,j}, \Sigma_{\theta,j}))) \\
             &= \min_j \left(  -\log(c g_j \mathcal{N}(\vec{\theta};\vec{\mu}_{\theta,j}, \Sigma_{\theta,j})) \right)
\end{align}
where $g_j$ are the mixture model weights of ${N=8}$ Gaussians, and $c$ a positive constant
required by our solver implementation. 
Although $E_\theta$ is not
differentiable at points where the mode with minimum energy changes,
we approximate its Jacobian by the Jacobian of the mode with minimum
energy in the current optimization step.

We define an interpenetration error term that exploits the capsule approximation
introduced in Sec.~\ref{subsec:capsules}. We relate the error term to the intersection volume between ``incompatible'' capsules (i.e.~capsules that do not intersect in natural poses).
Since the volume of capsule intersections is not simple to compute, we further simplify our capsules into spheres with centers $C(\vec{\theta},\vec{\beta})$
along the capsule axis and radius $r(\vec{\beta})$ corresponding to the capsule radius.
Our penalty term is inspired by the mixture of 3D Gaussians model in \cite{Sridhar:CVPR:2015}.
We consider a 3D isotropic Gaussian with $\sigma(\vec{\beta}) = \frac{r(\vec{\beta})}{3}$ for each sphere, and define the penalty as a scaled version of the
integral of the product of Gaussians corresponding to ``incompatible'' parts
\begin{align}
\label{eq:collisions}
E_{sp}(\vec{\theta}; \vec{\beta}) =
 \sum_{i} \sum_{j \in I(i)} \mathrm{exp}\left(\frac{||C_{i}(\vec{\theta},\vec{\beta})-C_{j}(\vec{\theta},\vec{\beta})||^2}{\sigma^2_{i}(\vec{\beta})+\sigma^2_{j}(\vec{\beta})}\right)
\end{align}
where the summation is over all spheres $i$ and $I(i)$ are the spheres incompatible with $i$.
Note that the term penalizes, but does not strictly avoid, interpenetrations. As desired, however, this term is differentiable with respect to pose and shape.
Note also that we do not use this term in optimizing shape since this would bias the body shape to be thin to avoid interpenetration.

We use a shape prior $E_\beta(\vec{\beta})$, defined as
\begin{align}
E_\beta(\vec{\beta}) = \vec{\beta}^T\Sigma^{-1}_\beta \vec{\beta}
\end{align}
where $\Sigma^{-1}_\beta$ is a diagonal matrix with the squared
  singular values estimated via Principal Component Analysis from the
  shapes in the SMPL training set. Note that the shape coefficients
  $\vec{\beta}$ are zero-mean by construction.
\subsection{Optimization}
\label{subsec:optimization}

We assume that camera translation and body orientation are unknown; we
require, however, that the camera focal length or its rough estimate
is known. We initialize the camera translation 
(equivalently $\vec{\gamma}$) by assuming that the
person is standing parallel to the image plane. Specifically, we
estimate the depth via the ratio of similar triangles, defined by the
torso length of the mean SMPL shape and the predicted 2D joints. 
Since this
assumption is not always true, we further refine this estimate by minimizing $E_J$ over the torso joints
alone with respect to camera translation and body orientation; we keep
$\vec{\beta}$ fixed to the mean shape during this optimization.
We do not optimize focal length, since the problem is too unconstrained to optimize it together with translation.

After estimating camera translation, we fit our model by minimizing
Eq.~(\ref{eq:objective}) in a staged approach.
We observed that starting with a high value for $\lambda_{\theta}$ and $\lambda_{\beta}$ and gradually decreasing them in the subsequent optimization stages is effective for avoiding local minima.

When the subject is captured in a side view, assessing in which direction the body is facing might be ambiguous.
To address this, we try two initializations when the 2D distance between the CNN-estimated 2D shoulder joints is below a threshold: first with body orientation
estimated as above and then with that orientation rotated by 180
degrees. Finally we pick the fit with lowest $E_J$.

We minimize Eq.~(\ref{eq:objective}) using Powell's dogleg method~\cite{nocedal}, using OpenDR and Chumpy~\cite{chumpy,loper14}. Optimization for a single image takes less than 1 minute on a common desktop machine.

\section{Evaluation}
\label{sec:experimental}
We evaluate the accuracy of both 3D pose and 3D shape estimation.
For quantitative evaluation of 3D pose, we use two
publicly available datasets: HumanEva-I \cite{HumanEva:2010} and Human3.6M \cite{Human36m:2014}.
We compare our approach to three state-of-the-art methods \cite{Akhter:2015,Ramakrishna:2012,Zhou:2015b} and also use these data for an ablation analysis.
Both of the ground truth datasets have restricted laboratory environments and limited poses.
Consequently, we perform a qualitative analysis on more challenging data from the Leeds Sports Dataset (LSP) \cite{LSP:2010}.
Evaluating shape quantitatively is harder since there are few images with ground truth 3D shape.
Therefore, we perform a quantitative evaluation using synthetic data to evaluate how well shape can be recovered from 2D joints corrupted by noise.
For all experiments, we use 10 body shape coefficients. 
We tune the $\lambda_{i}$ weights in Eq.~(\ref{eq:objective}) on the HumanEva training
data and use these values for all experiments.

\subsection{Quantitative Evaluation: Synthetic Data}
\label{subsec:data}

We sample synthetic bodies from the SMPL shape and pose space and project their joints into the image with a known camera. We generate 1000 images for male shapes and 1000 for female shapes, at $640 \times 480$ resolution.

\begin{figure}[t]
  \centering
\minipage{0.48\textwidth}
  \includegraphics[width=0.9\linewidth]{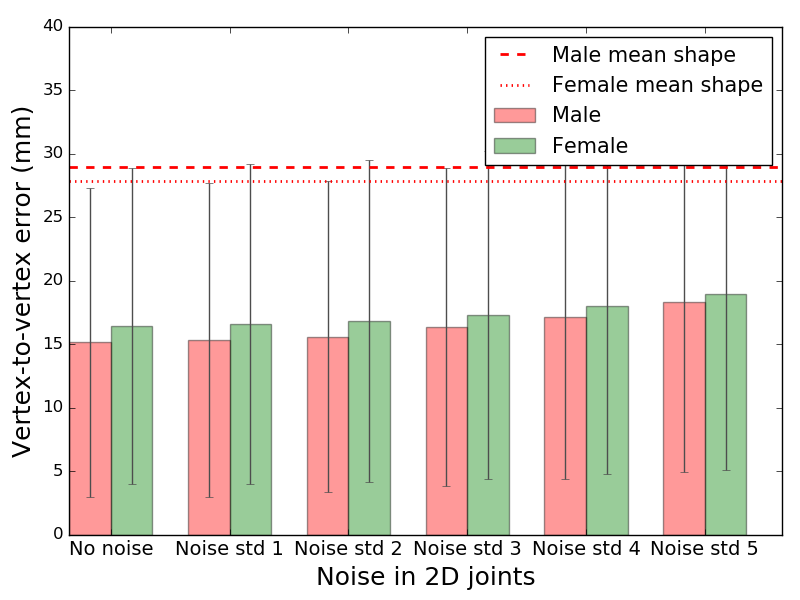}
\endminipage\hfill
\minipage{0.505\textwidth}
\includegraphics[trim={0 0 0
  1cm},clip,width=0.9\linewidth]{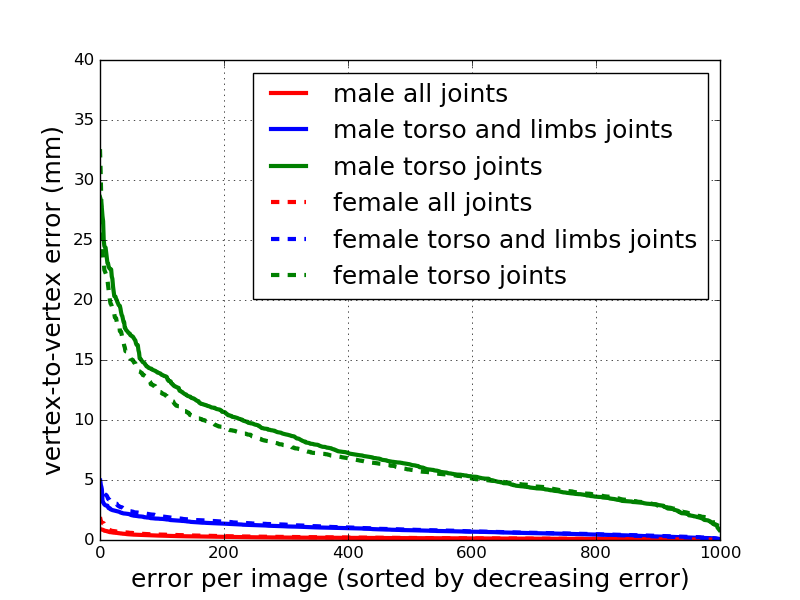}
\endminipage\hfill
\caption{{\bf Evaluation on synthetic data.} Left: Mean vertex-to-vertex
  Euclidean error between the estimated and true shape in a canonical pose, when
  Gaussian noise is added to 2D joints. Dashed and dotted lines represent the
  error obtained by guessing the mean shape for males and females,
  respectively. Right: Error between estimated and true shape when
  considering only a subset of joints during fitting.}\label{fig:noise}
\end{figure}

In the first experiment, we add varying amounts of i.i.d.~Gaussian
noise (standard deviation (std) from 1 to 5 pixels)
to each 2D joint. We solve for pose and shape by minimizing Eq.~(\ref{eq:objective}), setting the confidence weights for the joints in Eq.~(\ref{eq:joints}) to 1.
Figure~\ref{fig:noise} (left) shows the mean vertex-to-vertex Euclidean error between the estimated and true shape in a canonical pose.
Here we fit gender-specific models. The results of shape estimation are more accurate than simply guessing the average shape (red lines in the figure).
This shows that joints carry information about body shape that is relatively robust to noise.

In the second experiment, we assume that the pose is known, and try to
understand how many joints one needs to accurately
estimate body shape.
We fit SMPL to ground-truth 2D joints by minimizing
Eq.~(\ref{eq:joints}) with respect to: the full set of 23 SMPL joints;
the subset of 12 joints corresponding to torso and limbs (excluding head, spine, hands and feet); and the $4$ joints of the torso.
As above, we measure the mean Euclidean error between the estimated and true shape in a canonical pose.
Results are shown in Figure~\ref{fig:noise} (right).
The more joints we have, the better body shape is estimated.
To our knowledge, this is the first demonstration of estimating 3D
body shape from only 2D joints.
Of course some joints may be difficult to estimate reliably;
we evaluate on real data below.

\subsection{Quantitative Evaluation: Real Data}
\label{subsec:quantitative}

\paragraph{HumanEva-I.}
We evaluate pose estimation accuracy on single frames from the
HumanEva dataset \cite{HumanEva:2010}.
Following the standard procedure, we evaluate on the
Walking and Box sequences of subjects 1, 2, and 3 from the ``validation''
set \cite{Bo:IJCV10,Tekin:2015}.
We assume the gender is known and apply the gender-specific SMPL models.
 
Many methods train sequence-specific pose priors for HumanEva; we do
not do this.
We do, however, tune our weights on HumanEva training set and learn a
mapping from the SMPL joints to the 3D skeletal representation of HumanEva.
To that end we fit the SMPL model to the raw mocap marker data
in the training set using MoSh to estimate body shape and pose.
We then train a linear regressor from body vertices (equivalently
shape parameters $\vec{\beta}$) to the HumanEva 3D joints.
This is done once on training data for all subjects together and kept
fixed. We use the regressed 3D joints as our output for evaluation.

We compare our method against three state-of-the-art methods
\cite{Akhter:2015,Ramakrishna:2012,Zhou:2015b}, which, like us,
predict 3D pose from 2D joints. We report the average Euclidean distance between the ground-truth and predicted 3D joint
positions. Before computing the error we apply a similarity transform to align the
reconstructed 3D joints to a common frame via the Procrustes analysis
on every frame. 
Input to all methods is the same: 2D joints detected by DeepCut
\cite{Leonid2016DeepCut}.
Recall that DeepCut has not been trained on either dataset used for
quantitative evaluation. Note that these approaches have different skeletal structures of 3D
joints. We evaluate on the subset of 14 joints that semantically correspond
across all representations.
For this dataset we use the ground truth focal length.

Table \ref{table:heva} shows quantitative results where SMPLify achieves the lowest errors on all sequences.
While the recent method of Zhou et al. \cite{Zhou:2015b} is very good,
we argue that our approach is conceptually simpler and more accurate.
We simply fit the body model to the 2D data and let the model
constrain the solution. Not only does this ``lift'' the 2D joints to 3D, but SMPLify also produces a skinned vertex-based model that can
be immediately used in a variety of applications.

\begin{table}[b]
\centering
\begin{tabular}{lrrrrrrrr}
\hline
 & \multicolumn{3}{c}{Walking} & \multicolumn{3}{c}{Boxing} & Mean & Median \\
Method: & S1 & S2 & S3 & S1 & S2 & S3 &  \\
\hline
Akhter \& Black \cite{Akhter:2015} & 186.1 & 197.8 & 209.4 & 165.5 & 196.5 &
                                                                             208.4 & 194.4 & 171.2\\
Ramakrishna et al. \cite{Ramakrishna:2012} & 161.8 & 182.0 & 188.6 & 151.0 &
                                                                             170.4 & 158.3 & 168.4 & 145.9\\
Zhou et al. \cite{Zhou:2015b} & 100.0 & 98.89 & 123.1 & 112.5 & 118.6 & 110.0 &
                                                                                110.0
                       & 98.9 \\
SMPLify & \textbf{73.3} & \textbf{59.0} & \textbf{99.4} & \textbf{82.1} &
                                                                         \textbf{79.2} & \textbf{87.2} & \textbf{79.9}& \textbf{61.9} \\\hline
\end{tabular}
\caption{{\bf HumanEva-I results.} 3D joint errors in mm. }
\label{table:heva}

\end{table}

\begin{figure}[t]
\centerline{
\includegraphics[width=\linewidth]{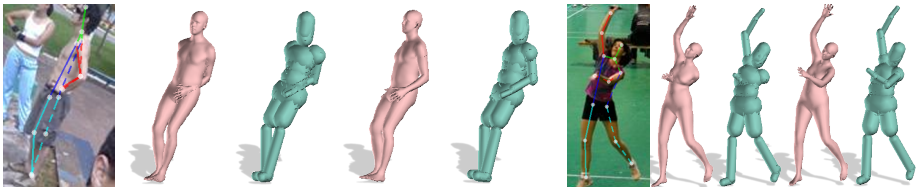}
}
\caption{{\bf Interpenetration error term.} Examples where the interpenetration
  term avoids unnatural poses. For each example we show, from left to right, CNN estimated joints, and the result of the optimization \emph{without} and \emph{with} interpenetration error term. }
\label{fig:selfpenetration}
\end{figure}

To gain insight about the method, we perform
an ablation study  (Table \ref{table:ablation}) where we evaluate
different pose priors and the interpenetration penalty term.
First we replace the mixture-model-based pose prior with $E_{\theta'}$, which uses a single Gaussian trained from the same data. 
This significantly degrades performance.
Next we add the interpenetration term, but this does not have a significant
impact on the 3D joint error. However, qualitatively, we find that it makes a difference in
more complex datasets with varied poses and viewing angles as illustrated in
Fig.~\ref{fig:selfpenetration}.

\begin{table}[b]
\centering
\begin{tabular}{llrrrrrrr}
\hline
 & \multicolumn{3}{c}{Walking} & \multicolumn{3}{c}{Boxing} &  Mean & Median\\
Method: & S1 & S2 & S3 & S1 & S2 & S3 & \\
\hline
$E_{\beta} + E_{J} + E_{\theta'}$
& 98.4 & 79.6 & 117.8 & 105.9 & 98.5 & 122.5 & 104.1 & 82.3 \\
$E_{\beta} + E_{J} + E_{\theta'}+E_{sp}$ & 97.9 & 79.4 & 116.0 & 105.8 & 98.5 &
                                                                                122.3
                  & 103.7 & 82.3 \\

SMPLify  & \textbf{73.3} & \textbf{59.0} & \textbf{99.4} & \textbf{82.1} &
                                                                         \textbf{79.2}
             & \textbf{87.2} & \textbf{79.9}  & \textbf{61.9}\\
\hline
\end{tabular}
\caption{{\bf HumanEva-I ablation study.} 3D joint errors in mm. The first row drops the
  interpenetration term and replaces the pose prior with a 
uni-modal prior. The second row keeps the uni-modal pose prior but adds the
interpenetration penalty. The third row shows the proposed SMPLify
model.}
\label{table:ablation}
\end{table}

\paragraph{Human3.6M.}
We perform the same analysis on the Human 3.6M dataset
\cite{Human36m:2014},  which has a wider range of poses.
Following \cite{Li:ACCV14,Tekin:2015,Zhou:CVPR:2016}, we report
results on sequences of subjects S9 and S11. 
We evaluate on five different action sequences captured
from the frontal camera (``cam3'') from trial 1. 
These sequences consist of 2000 frames on average and we evaluate on
all frames {\em individually}.
As above, we use training mocap and MoSh to train a regressor from the
SMPL body shape to the 3D joint representation used in the dataset. 
Other than this we do not use the training set in any manner.
We assume that the focal length as well as the distortion coefficients
are known since the subjects are closer to the borders of the image. 
Evaluation on Human3.6M is shown in Table \ref{table:H36} where our
method again achieves the
lowest average 3D error.
While not directly comparable, Ionescu et al.~\cite{Ionescu:CVPR:2014} report an error
of 92mm on this dataset.

\begin{table}[t]
\resizebox{\textwidth}{!}{
\begin{tabular}{lrrrrrrrrr}
\hline
                            & \multicolumn{1}{c}{Directions} & \multicolumn{1}{c}{Discussion} & \multicolumn{1}{c}{Eating}  & \multicolumn{1}{c}{Greeting} & \multicolumn{1}{c}{Phoning} & \multicolumn{1}{c}{Photo} & \multicolumn{1}{c}{Posing} & \multicolumn{1}{c}{Purchases} & \multicolumn{1}{c}{Sit}    \\ \hline
Akhter \& Black \cite{Akhter:2015}    & 199b.2                          & 177.6                          & 161.8                       & 197.8                        & 176.2                       & 186.5                     & 195.4                      & 167.3                         & 160.7                      \\
Ramakrishna et al. \cite{Ramakrishna:2012} & 137.4                          & 149.3                          & 141.6                       & 154.3                        & 157.7                       & 158.9                     & 141.8                      & 158.1                         & 168.6                      \\
Zhou et al. \cite{Zhou:2015b}       & 99.7                           & 95.8                           & 87.9                        & 116.8                        & 108.3                       & 107.3                     & 93.5                       & 95.3                          & 109.1                      \\
SMPLify                     & \textbf{62.0}                  & \textbf{60.2}                  & \textbf{67.8}               & \textbf{76.5}                & \textbf{92.1}               & \textbf{77.0}             & \textbf{73.0}              & \textbf{75.3}                 & \textbf{100.3}             \\ \hline
                            & \multicolumn{1}{c}{SitDown}    & \multicolumn{1}{c}{Smoking}    & \multicolumn{1}{c}{Waiting} & \multicolumn{1}{c}{WalkDog}  & \multicolumn{1}{c}{Walk}    & \multicolumn{2}{c}{WalkTogether}                       & \multicolumn{1}{c}{Mean}      & \multicolumn{1}{l}{Median} \\ \hline
Akhter \& Black \cite{Akhter:2015} & 173.7                          & 177.8                          & 181.9                       & 176.2                        & 198.6                       & \multicolumn{2}{r}{192.7}                              & 181.1                         & 158.1                      \\
Ramakrishna et al. \cite{Ramakrishna:2012} & 175.6                          & 160.4                          & 161.7                       & 150.0                        & 174.8                       & \multicolumn{2}{r}{150.2}                              & 157.3                         & 136.8                      \\
Zhou et al. \cite{Zhou:2015b} & 137.5                          & 106.0                          & 102.2                       & 106.5                        & 110.4                       & \multicolumn{2}{r}{115.2}                              & 106.7                         & 90.0                       \\
SMPLify                     & \textbf{137.3}                 & \textbf{83.4}                  & \textbf{77.3}               & \textbf{79.7}                & \textbf{86.8}               & \multicolumn{2}{r}{\textbf{81.7}}                      & \textbf{82.3}                 & \textbf{69.3}              \\ \hline
\end{tabular}
}
\caption{{\bf Human 3.6M.} {3D joint errors in mm}.
}
\label{table:H36}
\end{table}

\begin{figure}[t]
\centerline{
\includegraphics[width=\linewidth]{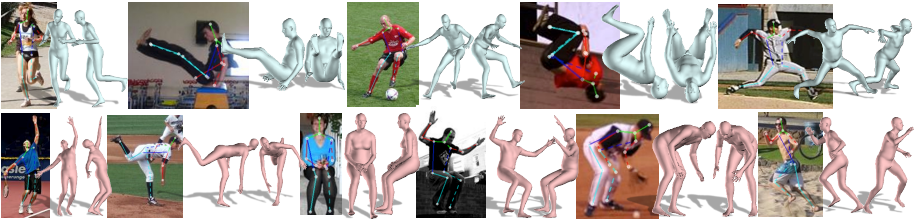}
}

\caption{{\bf Leeds Sports Dataset.}
Each sub-image shows the original image with the 2D joints fit by the CNN.
To the right of that is our estimated 3D pose and shape and the model seen from another view.
The top row shows examples using the gender-neutral body model; the bottom row show fits using the gender-specific models.}
\label{fig:LSPgood}
\end{figure}
\begin{figure}[b]
\centerline{
\includegraphics[width=\linewidth]{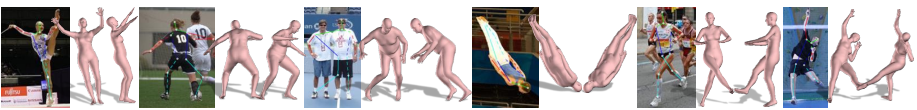}
}
\caption{{\bf LSP Failure cases.} Some representative failure
  cases: misplaced limbs, limbs matched with the limbs of other
  people, depth ambiguities.
}
\label{fig:LSPbad}
\end{figure}

\subsection{Qualitative Evaluation}
\label{subsec:qualitative}
Here we apply SMPLify to images from the Leeds Sports Pose (LSP) dataset
\cite{LSP:2010}.
These are much more complex in terms of pose, image resolution,
clothing, illumination, and background than HumanEva or Human3.6M.
The CNN, however, still does a good job of estimating the 2D poses.
We only show results on the LSP test set.
Figure~\ref{fig:LSPgood} shows several representative examples where
the system works well.
The figure shows results with both gender-neutral and gender-specific
SMPL models; the choice has little visual effect on pose.
For the gender-specific models, we manually label the images according
to gender.

\begin{figure}[t]
\centerline{
\includegraphics[width=\linewidth]{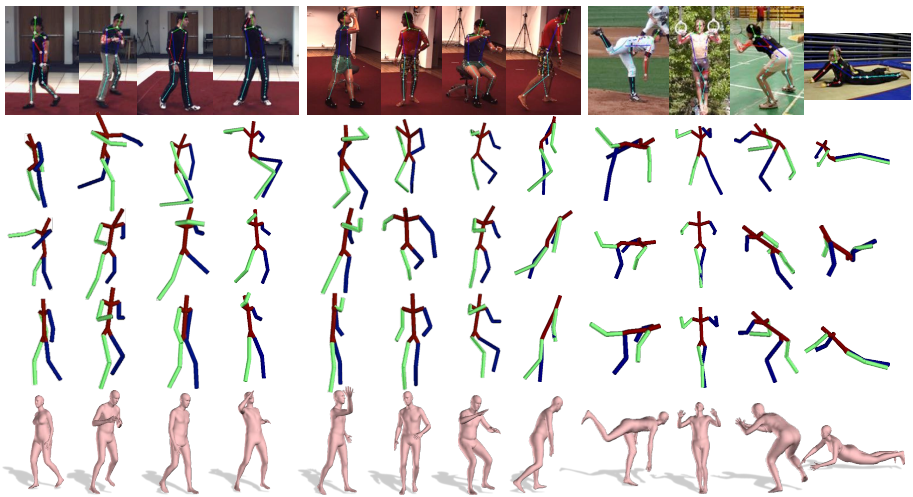}
}

\caption{{\bf Qualitative comparison.}
From top to bottom:
Input image.
Akhter \& Black \cite{Akhter:2015}.
Ramakrishna et al. \cite{Ramakrishna:2012}.
Zhou et al. \cite{Zhou:2015b}.
SMPLify.
}
\label{fig:Comparo}
\end{figure}

Figure~\ref{fig:Comparo} visually compares the results of the different methods on a few images from each of the datasets.
The other methods suffer from not having a strong model of how limb
lengths are correlated. LSP contains complex poses and these often show the value of the interpenetration term.
Figure~\ref{fig:selfpenetration} shows two illustrative
examples. Figure~\ref{fig:LSPbad} shows a few failure cases on LSP. 
Some of these result from CNN failures where limbs are mis-detected or are
matched with those of other people. Other failures are due to challenging
depth ambiguities. See Supplementary Material \cite{simplifyweb} for more results.

\section{Conclusions}
\label{sec:conclusions}
We have presented SMPLify, a fully automated method for estimating 3D body shape and pose from 2D joints in single images.
SMPLify uses a CNN to estimate 2D joint locations, and then fits a 3D human body model to these joints.
We use the recently proposed SMPL body model, which captures correlations in body shape, highly constraining the fitting process. 
We exploit this to define an objective function and optimize pose and shape directly by minimizing the error between the projected joints of the model and the estimated 2D joints. This gives a simple, yet very effective, solution to estimate 3D pose and approximate shape.
The resulting model can be immediately posed and animated.
We extensively evaluate our method on various datasets and find that SMPLify outperforms state-of-the-art methods.

Our formulation opens many directions for future work. 
In particular, body shape and pose can benefit from other cues such as silhouettes and we plan to extend the method to use multiple camera views and multiple frames.
Additionally a facial pose detector would improve head pose estimation and automatic gender detection would allow the use of the appropriate gender-specific model.
It would be useful to train CNNs to predict more than 2D joints, such as features related directly to 3D shape.
Our method provides approximate 3D meshes in correspondence with images, which could be useful for such training.
The method can be extended to deal with multiple people in an image; having 3D meshes should help with reasoning about occlusion.

\paragraph{Acknowledgements.}  We thank M.~Al Borno for inspiring the capsule representation, N.~Mahmood for help with the figures, I.~Akhter for helpful discussions.

\par\vfill\par

\clearpage

\bibliographystyle{splncs03}
\bibliography{egbib}
\end{document}